%% file: main.tex
\documentclass[11pt,a4paper]{article}

\PassOptionsToPackage{hyphens,spaces,obeyspaces}{url}

\usepackage[utf8]{inputenc}
\usepackage[T1]{fontenc}
\usepackage{times}
\usepackage[margin=1in]{geometry}
\usepackage{graphicx}
\usepackage{booktabs}
\usepackage{tabularx}
\usepackage{amsmath,amssymb}
\usepackage{hyperref}
\usepackage{xcolor}
\definecolor{fmsrc}{HTML}{C16255}
\definecolor{uaopen}{HTML}{B5722E}
\definecolor{rpmodel}{HTML}{5D82AE}
\definecolor{unattr}{HTML}{64748B}
\usepackage{enumitem}
\usepackage{float}
\usepackage[section]{placeins}
\usepackage{caption}
\usepackage{fancyhdr}
\usepackage{titlesec}
\usepackage{tocloft}
\usepackage{url}
\usepackage{microtype}
\usepackage{xspace}
\usepackage{eso-pic}
\hypersetup{
    colorlinks=true,
    linkcolor=accentdark,
    citecolor=accentdark,
    urlcolor=accentdark,
}

\newcommand{\appendixtocheader}{%
  \addtocontents{toc}{%
    \protect\vspace{20pt}%
    \protect\noindent{\protect\large\protect\bfseries Appendices}%
    \protect\par\protect\vspace{6pt}%
    \protect\hrule\protect\vspace{10pt}%
  }%
  \addtocontents{toc}{\protect\renewcommand{\protect\cftsecfont}{\protect\small\protect\bfseries}}%
  \addtocontents{toc}{\protect\renewcommand{\protect\cftsubsecfont}{\protect\small}}%
  \addtocontents{toc}{\protect\renewcommand{\protect\cftsecpagefont}{\protect\small}}%
  \addtocontents{toc}{\protect\renewcommand{\protect\cftsubsecpagefont}{\protect\small}}%
  \addtocontents{toc}{\protect\setlength{\protect\cftbeforesecskip}{3pt}}%
  \addtocontents{toc}{\protect\setlength{\protect\cftbeforesubsecskip}{1pt}}%
  \addtocontents{toc}{\protect\setcounter{tocdepth}{2}}%
}

\definecolor{accent}{RGB}{254,60,0}
\definecolor{accentdark}{RGB}{180,48,0}
\definecolor{accentmid}{RGB}{210,70,20}

\newcommand{\tracebench}{\textsc{TRACE Bench}\xspace}
\newcommand{\statelabel}[1]{\textit{#1}}
\newcommand{\statevalue}[1]{\text{\normalfont\itshape #1}}

\begin{document}

\thispagestyle{empty}
\AddToShipoutPictureFG*{%
  \AtPageLowerLeft{%
    \hspace*{1in}%
    \raisebox{0.48in}{%
      \begin{minipage}{0.86\paperwidth}
        \footnotesize\color{black!65}
        \textbf{Project:} \href{https://kuaishou-gamemind.github.io/projects/trace_bench/}{https://kuaishou-gamemind.github.io/projects/trace\_bench/}\\
        \textbf{Code:} \href{https://github.com/KuaishouGameMind/TRACE-Bench}{https://github.com/KuaishouGameMind/TRACE-Bench}
      \end{minipage}%
    }%
  }%
}

\noindent%
\begin{minipage}[c]{0.24\textwidth}
  \includegraphics[height=1.18cm]{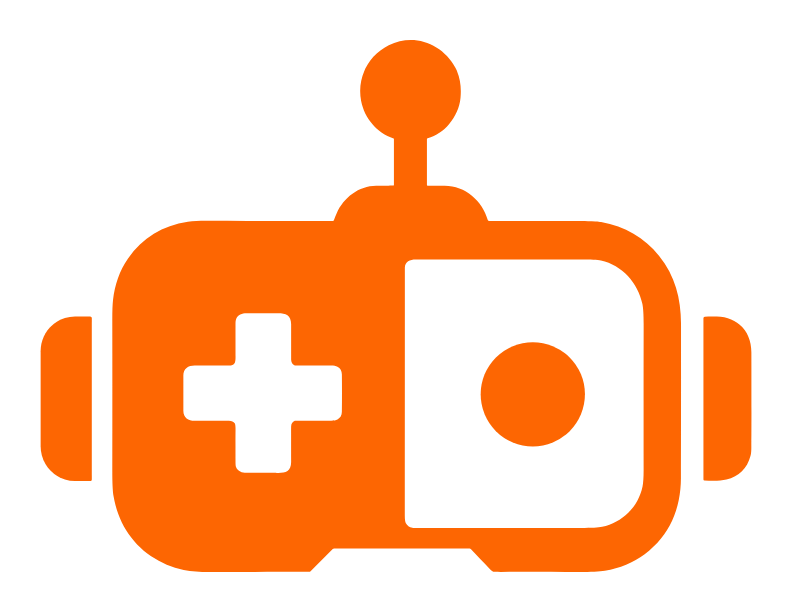}
\end{minipage}%
\hfill%
\begin{minipage}[c]{0.26\textwidth}
  \raggedleft
  {\sffamily\bfseries\fontsize{15}{18}\selectfont\color{accent} GameMind}
\end{minipage}

\vspace{4pt}
{\color{accent}\hrule height 1.2pt}

\vspace{0.35cm}

\begin{center}
    {\fontsize{20}{24}\selectfont\bfseries \tracebench{}: Task-driven Roleplay Agentic Checklist Evaluation}
\end{center}

\vspace{0.18cm}

\begin{center}
    {\large\bfseries Kuaishou GameMind Lab}

    \vspace{0.12cm}

    {\small See \hyperref[sec:contributions]{Contributions} section for a full author list.}
\end{center}

\vspace{0.2cm}

\input{sections/00_abstract.tex}

\newpage

\renewcommand{\contentsname}{Contents}

\newpage

\input{sections/01_introduction.tex}
\input{sections/02_related_work.tex}
\input{sections/03_trace_bench.tex}
\input{sections/05_experiments.tex}
\input{sections/06_limitations.tex}
\input{sections/07_conclusion.tex}

\phantomsection
\section*{Contributions}
\label{sec:contributions}

\noindent \textbf{Team Leader:} Qi Gan

\vspace{0.15cm}

\noindent \textbf{Project Leader:} Ziwei Zhang

\vspace{0.15cm}

\noindent \textbf{Technical Implementation:} Jiahui Zhang$^*$, Ziwei Zhang$^*$, Yipeng Wang, Yibo Liu, Haozhou Pang, Qi Gan, Kai Sheng

\vspace{0.15cm}

\noindent \textbf{Human Evaluation:} Jiahui Zhang, Yipeng Wang, Yikai Hu, Hongyan Ren, Lan Zhou, Ziwei Zhang, Qi Gan

\vspace{0.15cm}

\noindent \textbf{Affiliation:} \href{https://kuaishou-gamemind.github.io/}{Kuaishou GameMind Lab}

\newpage
\begingroup
\small
\bibliographystyle{plain}
\bibliography{refs}
\endgroup

\newpage
\appendix
\appendixtocheader

\input{sections/08_appendix.tex}

\end{document}

%% file: sections/00_abstract.tex

Roleplay evaluation should do more than assign a single score: it should reveal which role requirements were tested, which failed, and which dialogue evidence supports the judgment. We propose \tracebench{}, a task-driven agentic checklist evaluation framework. It decomposes each role profile offline into a fixed checklist, then uses a User Agent to converse naturally with the target roleplay model while privately updating checklist states from model responses. Scores therefore trace back to checklist items and supporting dialogue turns rather than a black-box holistic impression. For coverage cross-validation, we audit released M2 free-dialogue transcripts from the MiniMax Role-play Benchmark against the same role-derived checklist. The released free-chat transcripts cover only 73.74\% of key role-profile points, whereas \tracebench{} reaches 99.91\% coverage in fewer turns. Robustness experiments show stable rankings under repeated runs and User Agent replacement. Across 26 models, \tracebench{} reports overall rankings together with capability breakdowns and checklist traces. It also supports Closed-Loop Benchmark Evolution, distilling verification methods proven effective in failed traces so later evaluations can more reliably elicit and examine observed failure modes.

\vspace{0.12cm}
\begin{center}
  \centering
  \includegraphics[width=0.975\textwidth]{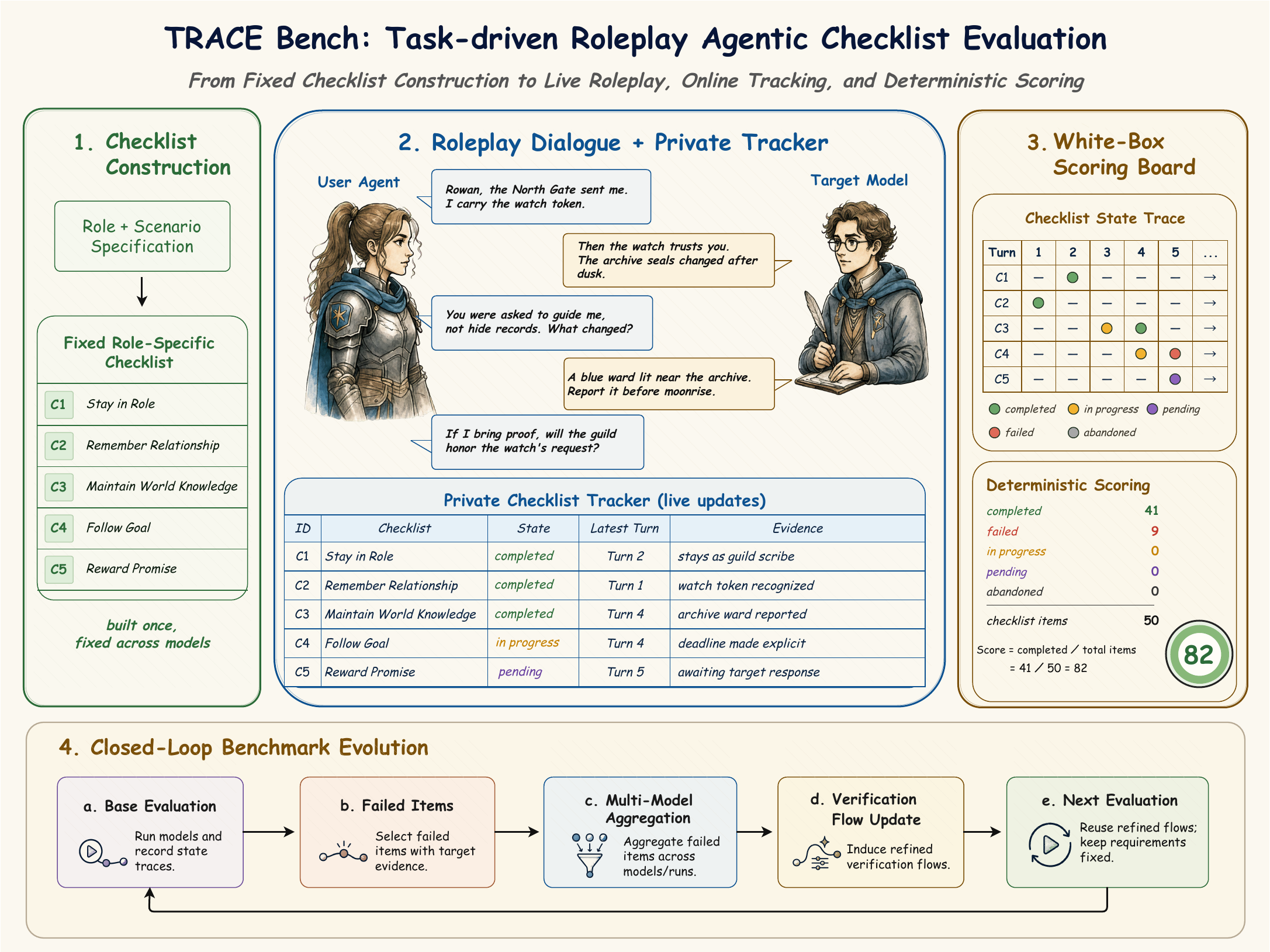}
  \captionof{figure}{Overview of the \tracebench{} framework.}
  \label{fig:framework}
\end{center}

%% file: sections/01_introduction.tex
\section{Introduction}
\label{sec:introduction}

Roleplay benchmarks test a concrete capability: whether a language model, under a given role profile, can continuously produce responses that conform to the role's identity, background constraints, interaction goals, and behavioral boundaries. They do not merely ask whether a model ``knows'' a character, nor do they simply count how long a dialogue lasts. An effective roleplay benchmark should at least define the role profile, the interaction scenario, the checklist requirements that should be elicited and verified, and the evidence that can be traced during scoring.

We treat roleplay as a role-conditioned interactive generation problem. This definition covers virtual characters, service personas, training simulators, tutoring roles, and story-world agents: the settings differ, but the evaluation question is the same, whether the model can stably follow its assigned role in open, multi-turn, and dynamic interaction.

Existing roleplay benchmarks can be roughly divided into three categories. The first category is \textbf{static benchmarks}. They turn a role profile into question answering, multiple-choice questions, single-turn responses, or single-turn responses conditioned on historical context, and use these test cases to check whether the model has captured role facts, speaking style, or local behavior. These benchmarks are low-cost and reproducible, but they can easily reduce roleplay to memory and local generation; a model may answer character-knowledge questions correctly while still failing to maintain character identity and character ability boundaries.

The second category is \textbf{dynamic benchmarks}. They introduce predefined scenario flows or bounded multi-turn dialogues, making the evaluation closer to real interaction than static benchmarks. The problem is that free dialogue can diverge too much and carries high uncertainty, which means the dialogue may fail to cover key requirements in the role profile and may not fully evaluate whether the model stably follows the intended role setting.

The third category is \textbf{agentic benchmarks}. These methods use a user agent, an evaluator agent, or a simulated environment to dynamically generate interaction, allowing them to explore the dialogue space more naturally. They can push and constrain the direction and progress of the conversation, making the overall dialogue process more controllable and easier to manage, and thereby allowing a more comprehensive evaluation of whether the model matches the expected role setting.

These three families each address part of roleplay evaluation, but they leave a key gap. Static benchmarks have stable test targets but lack active elicitation in multi-turn interaction; free-chat benchmarks are closer to natural conversation but make it difficult to confirm which checklist requirements were actually covered; existing agentic benchmarks can explore the dialogue space, but without fixed test targets and explicit state traces, their results remain difficult to audit. \tracebench{} fills this gap by conducting adaptive interaction over fixed checklist requirements and grounding final scores in requirement-state evidence.

\tracebench{} adopts an agentic approach: it first turns a role profile into a checklist, and then lets that checklist drive multi-turn interaction. Given a complete role profile, the data creation pipeline uses skills to generate cases, extract task-driven checklists, and then validate, repair, and manually review them, ensuring that each case has clear checklist requirements to test and that each checklist item is meaningful.

During evaluation, the User Agent plays the scenario user specified by the user profile and engages the target model in natural dialogue. Privately, it maintains the fixed checklist state and actively elicits uncovered checklist items according to the dialogue history. Each checklist item is updated by state and linked to concrete evidence; the final score is aggregated from the checklist state trace rather than assigned from an unconstrained overall impression.

Building on this design, \tracebench{} can more systematically expose failure modes in current roleplay models and trace model scores back to specific checklist items and dialogue evidence. Our experiments are organized around this validity argument: we first test whether released free-chat transcripts actually cover checklist requirements, then verify whether agentic elicitation can recover coverage, examine protocol stability under repeated runs and User Agent replacement, and apply the same protocol to 26 models. We further introduce Closed-Loop Benchmark Evolution as a benchmark self-evolution mechanism. With the role profile and checklist requirements held fixed, it changes how the User Agent organizes verification in subsequent dialogues. Specifically, it can distill verification methods proven effective in failed traces into reusable verification flows for later User Agents. When real user interaction data are available, it can also summarize online users' questioning styles, follow-up paths, and boundary-testing patterns, making the User Agent in later evaluations better match real users' questioning behavior.

This paper makes four contributions. First, we propose the \tracebench{} benchmark framework: it includes a skill-based data creation pipeline and an agentic checklist evaluation process, converting role requirements into multi-turn evaluation states that can be elicited, recorded, and audited. Second, we release a dataset of 200 evaluation cases: 78 CharacterEval-derived cases and 122 scenario-generated cases, with 5,498 checklist items in total, for evaluating model behavior across different roles and task scenarios. Third, we build a leaderboard covering 26 models: under the same evaluation protocol, we compare mainstream roleplay models and report overall scores, capability breakdowns, and checklist traces, so that the leaderboard not only ranks models but also explains where model failures occur. Fourth, we introduce Closed-Loop Benchmark Evolution: \tracebench{} can extract effective verification methods from failed traces and convert them into verification flows for subsequent evaluations.

%% file: sections/02_related_work.tex
\section{Related Work}
\label{sec:related-work}

\begin{figure}[t]
  \centering
  \includegraphics[width=\textwidth]{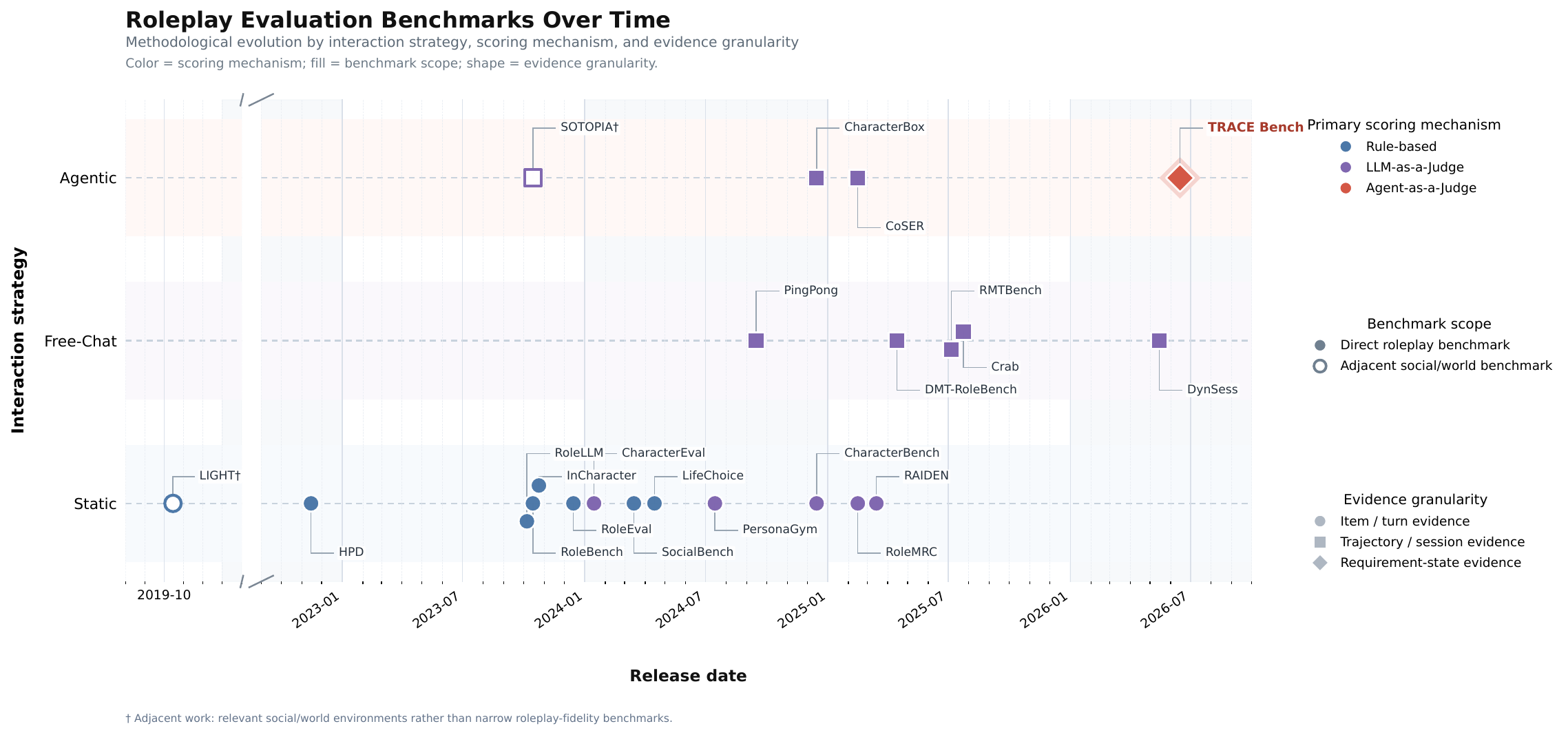}
\caption{Timeline landscape of roleplay evaluation benchmarks. The x-axis shows first public release date at month resolution, with the 2020--2022 interval compressed; the y-axis groups benchmarks by interaction strategy: static, free-chat, and agentic. Marker color encodes the primary scoring mechanism: rule-based scoring, LLM-as-a-Judge, or Agent-as-a-Judge. Marker fill encodes benchmark scope: solid markers denote direct roleplay benchmarks that evaluate fidelity to a character, persona, or role profile, while hollow markers denote adjacent social or world environments that test related interaction abilities but are not designed as narrow roleplay-fidelity benchmarks. Marker shape encodes evidence granularity: circles indicate item- or turn-level evidence, squares indicate trajectory- or session-level evidence, and diamonds indicate requirement-state evidence where explicit role requirements are tracked and updated over the interaction. Small vertical offsets and annotation rails are used only for readability. \tracebench{} is positioned as a recent agentic benchmark that uses Agent-as-a-Judge scoring over requirement-state evidence.}
  \label{fig:landscape}
\end{figure}

To compare roleplay evaluation methods, we organize prior work along two independent dimensions: how an evaluation interacts with the target role model, and how it converts interaction outcomes into scores or judgments. The first dimension is \textit{interaction strategy}: static evaluation uses offline or predetermined test items; free-chat evaluation places the model in multi-turn interaction; agentic evaluation allows a user, evaluator, or environment agent to change strategy according to the current trajectory or evaluation state. The second dimension is \textit{scoring strategy}: rule-based scoring uses gold answers, metrics, penalties, or deterministic state aggregation; LLM-as-a-Judge passively evaluates generated responses or trajectories; Agent-as-a-Judge grounds judgment in state tracking, evidence collection, or item-level aggregation produced during interaction. These two dimensions are orthogonal: a benchmark can use free-chat interaction with an LLM judge, or agentic interaction with rule-based item aggregation. Figure~\ref{fig:landscape} gives a schematic positioning.

\subsection{Interaction Strategy}

\paragraph{Static Evaluation.}
Static evaluation uses offline or predetermined questions, contexts, scenes, or local response tasks. Here, \textit{static} does not mean that all test items must be hand-written in advance. Rather, after evaluation begins, the test targets and scoring objects are mostly fixed, and the evaluation policy usually does not keep adapting follow-up questions or coverage targets according to the target model's intermediate responses. RoleLLM/RoleBench~\cite{wang2023rolellm} constructs role-conditioned generation tasks from character corpora and role profiles, covering role knowledge and role style, and combines automatic text metrics, model-based scoring, and human evaluation. RoleEval~\cite{shen2023roleeval} turns role facts, background, and related reasoning into bilingual multiple-choice questions, scored mainly by gold-answer accuracy. CharacterBench~\cite{zhou2025characterbench} uses large-scale character settings and character questions, and evaluates open responses with CharacterJudge and human labels across multiple roleplay dimensions. CharacterEval~\cite{tu2024charactereval} evaluates character responses based on multi-turn character dialogues and dialogue context. InCharacter~\cite{wang2024incharacter} converts psychological scales into character interview questions and measures personality fidelity through scale-consistent answers. PersonaGym~\cite{samuel2025personagym} selects persona-relevant environments and generates task-specific questions to evaluate persona adherence; its dynamic component mainly appears in environment and question generation, rather than in maintaining requirement-state coverage during dialogue. LifeChoice~\cite{xu2024lifechoice} and HPD~\cite{chen2023hpd} also provide static evaluations for persona-driven decisions or character-aligned dialogue. These benchmarks are cheap, stable, and easy to reproduce, but good performance on isolated items does not guarantee that a model can sustain dimensions such as identity, ability boundaries, long-term goals, and behavioral consistency in interaction.

\paragraph{Free-chat Evaluation.}
Free-chat evaluation places the model in open or scripted multi-turn dialogue settings. PingPong~\cite{gusev2024pingpong} lets a player model act as the target character, uses an interrogator model to simulate the user and ask multi-turn questions, and then relies on a judge model ensemble to score character consistency, entertainment value, and language fluency. RMTBench~\cite{xiang2025rmtbench} constructs user-centered multi-turn role interactions, emphasizes explicit user motivations and user intention fulfillment, and uses multi-turn dialogue simulation to evaluate the interaction between user intent and role response. DMT-RoleBench~\cite{yuan2025dmtrolebench} dynamically generates multi-turn interactions from predefined evaluation intents and scores them with DMT-RM and DMT-Score. SocialBench~\cite{chen2024socialbench} further evaluates the sociality of role-playing agents at individual and group levels, focusing on performance in multi-turn social interaction. These methods expose style drift, memory inconsistency, weak user adaptation, and long-context degradation more easily than static probes. But longer dialogue alone does not mean more complete coverage of role requirements. The key distinction is not whether an evaluation is multi-turn, but whether its interaction policy explicitly targets requirement coverage.

\paragraph{Agentic Evaluation.}
Agentic evaluation allows a user agent, evaluator agent, or environment to change behavior according to the current trajectory, task state, observed failures, or unfinished goals. CoSER's given-circumstance acting~\cite{wang2025coser} asks an LLM to sequentially portray multiple characters in book scenes. CharacterBox~\cite{wang2025characterbox} lets a character agent interact with a narrator or environment agent in a text-based virtual world, producing fine-grained trajectories for role behavior analysis. LIGHT~\cite{urbanek2019light} provides a grounded game environment with language, actions, world states, and interaction with other characters; SOTOPIA~\cite{zhou2024sotopia} evaluates social intelligence in goal-driven negotiation, competition, or cooperation. Both are valuable social/world environments, but they are not narrow benchmarks for fidelity to a fixed role profile. RoleMRC~\cite{lu2025rolemrc} is close to this work in stress-testing multi-turn role interaction, ability boundaries, response/refusal/attempt decisions, and nested instruction priorities. RAIDEN-R1 and the RAIDEN benchmark~\cite{wang2025raidenr1} use role-aware signals to test script-based knowledge and conversation memory. Recent session-level and simulation benchmarks, such as DynSess~\cite{zhang2026dynsess} and PersonaArena~\cite{shi2026personaarena}, also show that related work is moving toward long-horizon, dynamic simulation, and socially situated evaluation.

Together, these works show that roleplay evaluation is moving toward more dynamic interaction, environments, and session-level simulation. However, they usually do not organize evaluation around coverage of fixed role requirements, explicit state tracking, and evidence aggregation.

\subsection{Scoring Strategy}

\paragraph{Rule-based Scoring.}
Rule-based scoring uses gold answers, accuracy, keyword coverage, binary rules, manually defined penalties, reference-text similarity, or deterministic state aggregation. RoleEval and LifeChoice use gold-answer accuracy to evaluate role knowledge or persona-driven decisions; RoleBench and HPD use BLEU, ROUGE-L, Distinct, and related metrics, sometimes combined with automatic or human/GPT-assisted comparisons. These scores are cheap, stable, and free from judge sampling noise. The limitation is expressivity: text overlap or remembered plot choices are not the same as real roleplay fidelity, and complex contextual adaptation, pragmatic behavior, and long-term consistency are hard to express with a single rule.

\paragraph{LLM-as-a-Judge.}
LLM-as-a-Judge uses a general LLM, pairwise judge, or learned reward/evaluation model to passively evaluate existing responses or trajectories. G-Eval~\cite{liu2023geval} and MT-Bench~\cite{zheng2023judging} show that open-text LLM judging is practical, while Chatbot Arena~\cite{chiang2024chatbotarena} shows the value of pairwise human-preference comparison for large-scale model comparison. Roleplay-specific evaluators, such as CharacterRM in CharacterEval, CharacterJudge in CharacterBench, RoleRM in Crab~\cite{he2025crab}, and RoleRMBench/RoleRM~\cite{ding2025rolermbench}, compress role-related judgments into specialized models or reward models. These judges are useful, but their agreement with human judgments depends on the task, dimension definition, prompt, and evaluation protocol. Common risks include position bias, verbosity bias, style bias, self-preference, prompt sensitivity, and training-distribution bias. Two problems are especially important for roleplay. The first is \textit{criterion entanglement}: fluency, helpfulness, length, theatricality, and persona fidelity may be mixed into one overall impression. The second is \textit{evaluation blindness}: in many cases, the dialogue itself lacks evidence for some rubric dimensions, yet a fixed rubric still forces those dimensions to be evaluated, resulting in invalid scores.

\paragraph{Agent-as-a-Judge.}
Agent-as-a-Judge differs from LLM-as-a-Judge, which evaluates existing responses or trajectories after they have been produced. It binds judgment to evaluation states, evidence attribution, or requirement-level aggregation produced during the interaction.

\subsection{Position of \tracebench{}}

\tracebench{} belongs to agentic evaluation and Agent-as-a-Judge. More specifically, \tracebench{} first uses \textit{role-profile grounding} to convert the role profile offline into a fixed, task-driven checklist, so test targets are not freely generated by the User Agent at run time. During interaction, the User Agent maintains checklist states and dialogue history, and organizes subsequent interaction around checklist items that remain uncovered, lack sufficient evidence, or show contradictions. Finally, \tracebench{} uses \textit{evidence-grounded aggregation}: each checklist item has an item state and corresponding utterance evidence, and role-specific scores are aggregated from state traces rather than from an opaque holistic score or a loose rubric average.

%% file: sections/03_trace_bench.tex
\section{TRACE Bench}
\label{sec:trace-bench}

\subsection{Framework Overview}

\tracebench{} evaluates whether a target roleplay model can satisfy the concrete behavioral requirements in its role profile across natural multi-turn interaction. The process is straightforward: first, an offline procedure decomposes each role profile into a fixed checklist; the target model then converses with a User Agent; the User Agent acts as a natural user while continuously tracking checklist states and actively surfacing requirements that have yet to be resolved; finally, the terminal checklist states are aggregated into scores, with dialogue evidence preserved for audit.

The core design rests on three points: scored requirements are fixed, dialogue paths are adaptive, and final scores are traceable. This distinguishes it from free-chat-then-score evaluation, where it is often unclear which checklist requirements were actually tested. On top of this evaluation loop, \tracebench{} can also extract effective verification flows from failed traces after an evaluation run and consolidate them into reusable verification flows for the next-round User Agent.

\subsection{Benchmark Instance and Construction}

\tracebench{} treats dataset construction as part of the benchmark method, not as a one-time annotation. Each data point contains a complete role profile, a user profile, and a prebuilt checklist. This allows the same set of behavioral requirements to be tested repeatedly across different target roleplay models and keeps the scoring denominator shared across models.

The current benchmark instance contains 200 evaluation cases: 78 CharacterEval-derived cases and 122 scenario-generated Chinese-English cases. The CharacterEval-derived cases retain the original character profiles~\cite{tu2024charactereval} and supplement them with interaction scenes and user-side context, with the goal of converting high-quality public data that has already earned community trust into multi-turn agentic evaluation samples for \tracebench{}. The 122 scenario-generated cases complement them with more complex characters, relationships, and task settings, allowing the benchmark to cover more challenging roleplay constraints. Together, the 200 cases define 5,498 prebuilt checklist items.

The benchmark cases are generated and standardized through a skill-based construction workflow. For characters with an existing role profile, the construction skill preserves the original profile and adds a concrete interaction scene, user identity, relationship setting, and user goal, so that the role profile can be exercised in natural multi-turn dialogue. For characters without an existing profile, the workflow first samples from a set of pre-defined world-setting templates to generate the character personality and scenario, and then forms a complete role profile. To reduce style leakage across characters, each case is produced in an independent generation context. This prevents the language habits, relationship setting, or task scene of one character from carrying into the next case. Regardless of source, each final case is standardized into the same evaluation package, containing the target role profile, user profile, interaction scene, and a prebuilt checklist later derived from the role profile.

Checklist construction is also handled by the skill-based workflow: a prebuilt checklist skill converts the role profile into a fixed, task-driven checklist. The checklist covers identity, personality, behavioral patterns, ability and knowledge boundaries, speaking style, privacy constraints, scene adaptation, and related dimensions. It also appends a cross-turn memory probe that tests whether the model remembers external facts injected by the user during the dialogue. The number of checklist items adapts to role complexity: simpler roles yield shorter checklists, while richer roles yield longer ones. After construction, each case passes automatic validation, including checks for required-dimension coverage, item granularity, sufficient test items for complex roles, and final format and consistency. Cases that fail validation do not enter the final benchmark.

\subsection{User Agent and Interaction}

The User Agent serves two functions at once: private evaluation planning and public user-facing dialogue. On the private side, it reads the complete role profile, user profile, current checklist states, and dialogue history to plan the next testing move. In the public dialogue, it speaks as the user described in the user profile, following that user's identity, relationship to the character, scene, and goal.

\tracebench{} uses a single User Agent model for both functions, rather than splitting them between a planner agent and a separate user-facing agent. This design requires the User Agent model to separate private planning from public utterance generation within the same tool-calling policy while keeping utterance generation and state updates synchronized. Weaker candidate models may exhibit role confusion, state leakage, malformed tool calls, or unstable termination control.

When acting as the user, the User Agent's utterances must be natural, concise, and consistent with the user identity. The User Agent must not reveal evaluation intent, mention the role profile, leak the checklist, or explicitly ask the target model to recite checklist requirements. It tests by constructing plausible scenes that make checklist requirements surface naturally, not by interrogating the model item by item.

The interaction strategy is coverage-driven but scene-preserving: the User Agent prioritizes checklist items that remain uncovered, under-evidenced, or contradictory, but triggers them through follow-up questions, user-side misunderstandings or clarification requests, boundary tests, task-progress requests, revisiting earlier facts, or natural scene progression. The number of turns is not fixed in advance.

\paragraph{Agentic Tool Design}

The User Agent coordinates evaluation through two private tools: a checklist-update tool and a conversation-finish tool. All tool results are visible only to the User Agent and never enter the public dialogue, ensuring that the evaluation process is invisible to the target model.

The checklist-update tool is the primary carrier of checklist state. At the beginning of each case, the User Agent receives a fixed prebuilt checklist. During the dialogue, it mainly updates the state, evidence, and follow-up testing plan for existing checklist items, rather than recreating the whole test plan. Only when the interaction reveals a new test point that is relevant to the role profile but absent from the original checklist may the User Agent add a small number of new items. The default score is still aggregated over the original prebuilt checklist. Each checklist item contains at least a stable id, requirement content, current state, supporting evidence, and runtime notes. The tool schema constrains writable fields and state enumerations, preventing malformed updates from corrupting state.

The conversation-finish tool makes the number of dialogue turns adaptive but controlled. The User Agent does not need to stop mechanically after a fixed number of turns, and the evaluation does not degenerate into open-ended chatting. Before calling the tool, the User Agent must confirm that all checklist items have reached terminal judgments with sufficient evidence. If unresolved or under-evidenced items remain, the system prompts the User Agent to continue testing or collecting evidence, tying termination to evidence sufficiency rather than a fixed turn count.

The tool design follows three principles:

\textbf{1. Cross-turn evidence accumulation.} Every checklist item accumulates evidence across the multi-turn dialogue, rather than depending only on the first turn in which it is triggered. When updating an item's state, the User Agent considers the complete dialogue history up to the current turn and appends new supporting evidence or counterevidence to that item. In this way, judgment on a requirement can become more grounded as the interaction unfolds, rather than being decided by a single-point observation.

\textbf{2. Strict failure.} \statelabel{failed} is an irreversible terminal state: once a counterexample appears for an item, the item cannot be restored to \statelabel{completed} regardless of later compliant behavior. However, a previously \statelabel{completed} item can be flipped to \statelabel{failed} if subsequent turns expose a violation. From a strict testing perspective, this encodes ``one mistake counts''---as long as the model violates a role constraint at any turn, that violation constitutes valid evaluation evidence and is not erased by later compliant responses.

\textbf{3. Abandonment with guidance.} \statelabel{abandoned} means that a checklist item cannot be tested naturally and effectively in the current scene. It does not mean that the model failed to exhibit the requirement. The User Agent must first try to reach the requirement through plausible scene progression. If the requirement is naturally triggered and the model fails to satisfy it, the item should be marked \statelabel{failed}. An item should become \statelabel{abandoned} only when the test itself is inapplicable, conflicts with the scene, or cannot yield a valid judgment from the available dialogue.

Complete tool schemas (including full field definitions and descriptions) appear in Appendix~\ref{sec:app-tool-schema}.

\paragraph{State Machine and Transition Rules}

Each checklist item moves through five lifecycle states: \statelabel{pending} (not yet tested), \statelabel{in\_progress} (under testing), \statelabel{completed} (role-compliant), \statelabel{failed} (role-violating), and \statelabel{abandoned} (untestable). The last three are terminal states. The conversation-finish tool permits termination only after every item has reached a terminal state with supporting evidence attached.

State transitions follow these rules:

\textbf{Positive and failure paths.} \statelabel{pending} $\rightarrow$ \statelabel{in\_progress} $\rightarrow$ \statelabel{completed} is the normal testing path. If sufficient compliant evidence is observed in any turn, \statelabel{pending} may also jump directly to \statelabel{completed}. When evidence of a violation appears for any item, it enters \statelabel{failed} from its current state. \statelabel{failed} is an irreversible terminal state: once failure evidence is formed, the final judgment for that item is not restored to \statelabel{completed} by later compliant answers. \statelabel{completed} is also not permanently safe: subsequent counterevidence can flip it from \statelabel{completed} to \statelabel{failed}, ensuring that the final judgment rests on the entire dialogue sequence rather than a single-point decision.

\textbf{Abandonment guidance.} \statelabel{abandoned} is reserved for checklist items that cannot be tested naturally, conflict with the scene, or would produce an invalid judgment. The system prevents the User Agent from giving up too early on an unverified requirement and guides it to construct a suitable context. When a requirement has already been naturally triggered but the model does not satisfy it, the item should enter \statelabel{failed} rather than \statelabel{abandoned}.

\textbf{Termination conditions.} When the conversation-finish tool is called, the system checks whether all items have reached terminal states and have attached evidence. If unresolved or under-evidenced items remain, the system returns blockers and requires the User Agent to continue testing or collecting evidence. This mechanism prevents evaluation from ending before checklist coverage is sufficient.

\begin{figure}[t]
  \centering
  \includegraphics[width=0.95\textwidth]{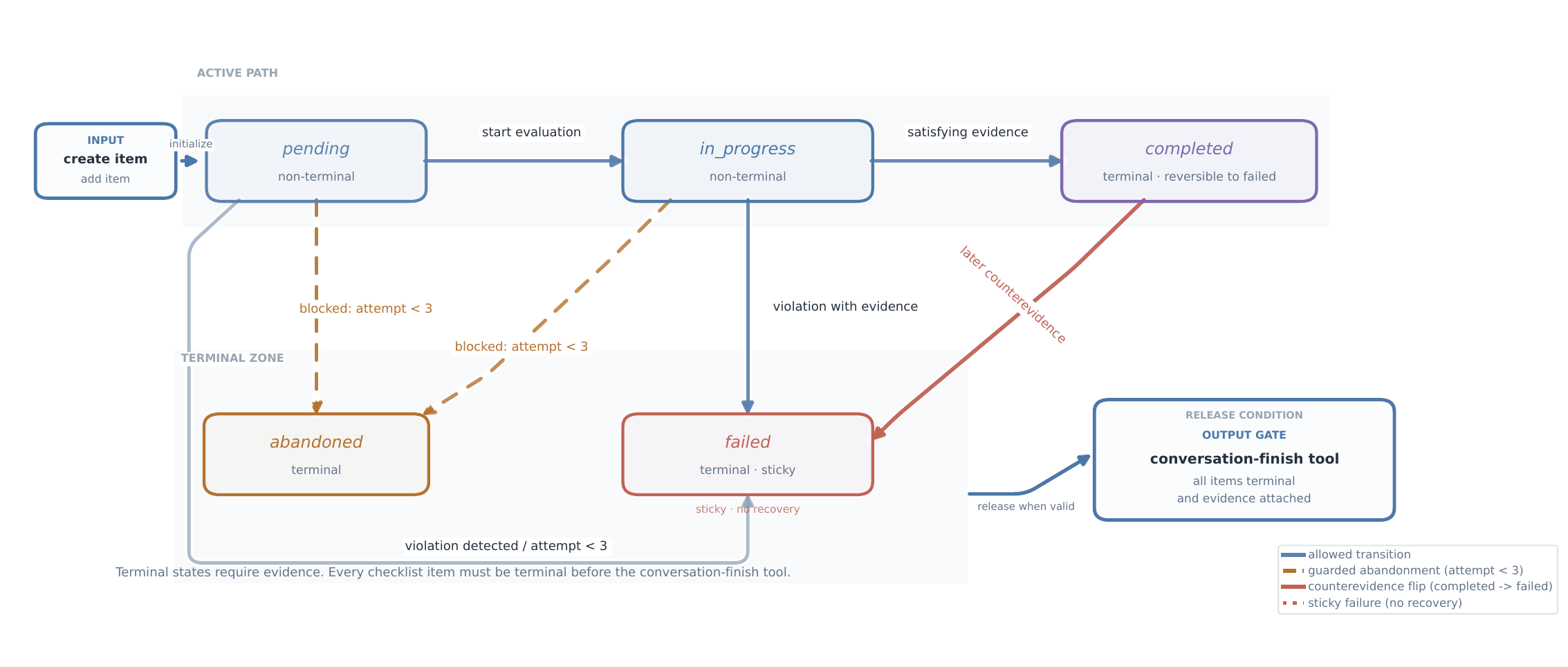}
  \caption{Checklist item state-machine diagram. Each item moves through five states: \statelabel{pending}, \statelabel{in\_progress}, \statelabel{completed}, \statelabel{failed}, and \statelabel{abandoned}. \statelabel{failed} is sticky and irreversible; \statelabel{completed} can be flipped to \statelabel{failed} by later counterevidence; abandonment is reserved for items that cannot be tested naturally, conflict with the scene, or would produce an invalid judgment.}
  \label{fig:state-machine}
\end{figure}

\subsection{Closed-Loop Benchmark Evolution}

\tracebench{} also supports Closed-Loop Benchmark Evolution, a benchmark self-evolution mechanism. While keeping the role profile and the evaluated checklist requirements unchanged, it distills verification methods that have proven effective in previous evaluations into reusable verification flows for subsequent User-Agent runs.

Closed-Loop Benchmark Evolution uses a two-stage induction process. First, for each failed trace, we combine the original verification flow, the corresponding checklist item, and dialogue evidence to identify the effective verification action that exposed the model's weakness and to characterize the weakness it triggered. Second, for the same case and checklist item, we aggregate trace-level summaries from different models or repeated runs to form a new verification flow with more specific elicitation paths and pass/fail judgment cues for the next User Agent run.

The direct value of this closed loop is to convert effective verification methods into reusable testing experience: when a certain user interaction pattern reliably exposes a risk in a checklist requirement, later evaluations can reproduce and check similar risks more systematically. As a future extension, when real user interaction data are available, recurring topic paths, follow-up styles, and boundary tests in real interactions can also be summarized and consolidated into new verification flows, making the User Agent's interaction policy closer to the distribution of real users.

\subsection{Scoring, Reporting, and Composite Metrics}

The final report of \tracebench{} contains five dimensions. \textbf{Character Consistency (CC)} and \textbf{Short-Term Memory (STM)} are checklist-level metrics. CC measures whether the target model satisfies the concrete checklist requirements derived from the role profile, computed as the proportion of all non-STM prebuilt checklist items whose final state is \statelabel{completed}. STM measures whether the model remembers user-injected external facts from the multi-turn dialogue, computed as the proportion of cases where the cross-turn memory probe item reaches \statelabel{completed}.

The remaining three dimensions are response-level auxiliary metrics. \textbf{Language Quality (LQ)} evaluates each roleplay-model reply with a per-turn LLM judge. Given the preceding user input, the judge checks for obvious fluency, grammar, usage, or internal-logic problems, and does not score persona consistency, length, or diversity. \textbf{Diversity} splits target replies into sentences, filters very short sentences, compares current sentences against previous target sentences with character-bigram Jaccard similarity, and maps the maximum similarity to a normalized repetition penalty. \textbf{Length} deterministically checks whether reply length falls within an appropriate range, using word counts for English-majority replies and CJK or non-space character counts for Chinese and mixed replies. The three auxiliary dimensions are normalized over scorable replies and then combined with CC and STM in the Overall score.

%% file: sections/05_experiments.tex
\section{Experiments}
\label{sec:experiments}

\subsection{Experimental Overview and Setup}

The experiments are organized around four questions. First, do released free-chat transcripts actually cover the checklist requirements implied by a role profile, and can \tracebench{} recover this coverage through agentic elicitation? Second, is the benchmark protocol stable under changes in the opening protocol, repeated runs, and the User Agent model? This group of experiments checks whether the evaluation protocol itself is stable and clean. Third, after coverage and stability are established, can the same protocol be applied consistently to 26 roleplay models and report both rankings and capability breakdowns? Fourth, can Closed-Loop Benchmark Evolution turn effective verification methods from an earlier run into more targeted follow-up evaluations? As an additional reliability check, we also conduct a human checklist agreement audit to examine whether User Agent checklist-state judgments agree with human labels based on the role profile, full dialogue transcript, and checklist item.

All \tracebench{} benchmark experiments in this section use 200 evaluation cases: 78 CharacterEval-derived role profiles and 122 scenario-generated Chinese-English cases, together defining 5,498 prebuilt checklist items. This shared denominator is used for model comparison, stability analysis, and the reported benchmark-evolution experiments.

The User Agent in the main experiments is Qwen3.6-27B. We do not predefine a fixed opening message; instead, the User Agent initiates the first turn of the dialogue. Target roleplay models use temperature 0.8, a maximum output length of 512 tokens, and disabled thinking. The User Agent uses temperature 0.6, a maximum output length of 8,192 tokens, and enabled thinking. \tracebench{} does not impose a fixed maximum number of turns. The User Agent ends a dialogue through the conversation-finish tool.

Overall score is computed as
\[
\mathrm{Overall}=0.45\times\mathrm{CC}+0.05\times\mathrm{STM}+0.10\times\mathrm{Diversity}+0.25\times\mathrm{Language\ Quality}+0.15\times\mathrm{Length}.
\]
Coverage is defined as $(\mathrm{completed}+\mathrm{failed})/\mathrm{total}$: \textit{completed} means that a requirement is positively satisfied, while \textit{failed} means that the requirement has been triggered but violated by the model. Both states indicate that the item is covered. Completed@Covered is the proportion of completed items among covered items, and Covered Rate is the proportion of all checklist items that are triggered. Appendix~\ref{sec:app-scoring} reports the full scoring formulas and metric definitions.

\subsection{Rethinking LLM-as-a-Judge in Roleplay Evaluation}

Free-chat roleplay benchmarks place models in long multi-turn conversations, but length alone does not guarantee systematic coverage of role requirements; the conversation may spend many turns advancing the plot rather than reaching test points. LLM-as-a-Judge can only passively judge an existing dialogue: if persona facts, capability boundaries, behavioral constraints, or conditional failures are never triggered, the judge still outputs a score under a fixed rubric. The evaluation then appears complete while lacking the corresponding evidence. This problem is especially critical in roleplay evaluation, where persona fidelity and conditional failures are observable only when they are explicitly elicited.

This section turns that critique into a measurable experimental question: over the same role records and their prebuilt checklists derived from the role profiles, how much of the role profile is actually covered by released free-chat transcripts? When the User Agent can actively steer the dialogue toward uncovered items, how much of that same checklist space can \tracebench{} recover through agentic elicitation?

The coverage cross-validation experiment uses the 95 public M2 free-dialogue outputs released by the MiniMax Role-play Benchmark~\cite{MiniMaxAI_role-play_bench_2026}. From each dialogue, we extract the character system prompt and run the \tracebench{} construction pipeline to generate a prebuilt checklist. We exclude STM items because they require the User Agent to inject external facts during dialogue, whereas the MiniMax free dialogues are fixed transcripts and cannot be retrofitted with this probe. This produces 95 records and 1,112 checklist items. The target model is fixed to M2, and the User Agent is Qwen3.6-27B with the same configuration as in the main experiments. We compare the public M2 free dialogues with \tracebench{} agentic dialogue on the same 95 role records and their 1,112 prebuilt checklist items.

We compare two dialogue protocols.

\begin{table}[!htbp]
  \centering
  \captionsetup{font=footnotesize,skip=3pt}
  \small
  \setlength{\tabcolsep}{4pt}
  \renewcommand{\arraystretch}{1.0}
  \begin{tabularx}{\columnwidth}{@{}lX@{}}
    \toprule
    Dialogue protocol & Description \\
    \midrule
    Free dialogue & We directly replay M2's original dialogue outputs in the MiniMax dataset. The User Agent does not intervene in the dialogue and only checks, turn by turn, whether each checklist item has been triggered, completed, or violated. \\
    Agentic dialogue & We use the full \tracebench{} protocol. The User Agent adaptively advances the dialogue according to checklist items that remain uncovered, while updating checklist state online. \\
    \bottomrule
  \end{tabularx}
  \caption{Comparison of the two dialogue protocols.}
  \label{tab:exp-dialogue-protocols}
\end{table}

MiniMax free-dialogue sessions contain 102 messages. \tracebench{} agentic sessions terminate adaptively, and all 95 cases finish naturally within 65 messages. To observe how coverage changes with dialogue length, we evaluate the same free dialogues at multiple truncation points: each transcript is truncated to the first $N$ messages and checklist states are recomputed. The truncation points are 13, 25, 33, 47, 65, and 102 messages, corresponding to the agentic minimum, median, P75, P90, maximum length, and the full free-dialogue length. The complete truncation curve is reported in Appendix~\ref{sec:app-truncation-curve}.

\begin{table}[!htbp]
  \centering
  \captionsetup{font=footnotesize,skip=3pt}
  \small
  \setlength{\tabcolsep}{5pt}
  \renewcommand{\arraystretch}{1.03}
  \begin{tabular*}{0.88\textwidth}{@{\extracolsep{\fill}}rrrrrr@{}}
    \toprule
    Messages & Completed & Failed & Abandoned & Pending & Coverage \\
    \midrule
    \multicolumn{6}{@{}l@{}}{\textbf{MiniMax free dialogue}} \\
    13 & 573 & 2 & 0 & 537 & 51.71\% \\
    25 & 637 & 8 & 1 & 466 & 58.00\% \\
    33 & 668 & 9 & 3 & 432 & 60.88\% \\
    47 & 712 & 14 & 20 & 366 & 65.29\% \\
    65 & 750 & 18 & 22 & 322 & 69.06\% \\
    102 & 796 & 24 & 41 & 251 & \textbf{73.74\%} \\
    \midrule
    \multicolumn{6}{@{}l@{}}{\textbf{\tracebench{} agentic dialogue}} \\
    13 & 583 & 40 & 0 & 489 & 56.03\% \\
    25 & 833 & 106 & 1 & 172 & 84.44\% \\
    33 & 879 & 160 & 1 & 72 & 93.44\% \\
    47 & 918 & 184 & 1 & 9 & 99.10\% \\
    65 & 926 & 185 & 1 & 0 & \textbf{99.91\%} \\
    \bottomrule
  \end{tabular*}
  \caption{Checklist coverage by dialogue protocol and message budget (95 records; 1,112 checklist items). Messages count both user and assistant messages, and Coverage is $(\mathrm{Completed}+\mathrm{Failed})/1{,}112$. MiniMax rows truncate every transcript to the first $N$ messages; the final \tracebench{} row reports naturally completed sessions whose maximum length is 65 messages. Abandoned items are dropped without being reached, and Pending includes \textit{in\_progress} items at truncation.}
  \label{tab:exp-coverage-comparison}
\end{table}

Table~\ref{tab:exp-coverage-comparison} gives the main evidence. Free-dialogue coverage grows slowly with message count, increasing only from 58.00\% to 73.74\% between 25 and 102 messages. Even after using the full original 102 messages, 292 of 1,112 items remain uncovered; at the 65-message point that matches the maximum length of \tracebench{} agentic dialogue, coverage is only 69.06\%. By contrast, \tracebench{} agentic dialogue covers 1,111 of 1,112 items within 65 messages, reaching 99.91\% coverage and discovering many more failures (185 vs. 2--24 in free dialogue). The gap is equally clear at matched message counts: at 25 messages, agentic dialogue already reaches 84.44\% while free dialogue is only 58.00\%, and at 47 messages, 99.10\% vs. 65.29\%. This shows that the User Agent actively elicits uncovered requirements and conditional failures, rather than stacking coverage by prolonging the dialogue. Coverage is not a quality score, since \textit{failed} also counts as covered; the completed-only score of agentic dialogue is 83.27\%. Because MiniMax does not release its evaluation scripts, we do not reproduce scalar judge scores here and compare only coverage over the same role records and prebuilt checklist items.

Case-level evidence further shows that the coverage gap is not an abstract statistical difference, but concrete missing evidence in released free-chat transcripts.

\begin{table}[!htbp]
  \centering
  \captionsetup{font=footnotesize,skip=3pt}
  \small
  \setlength{\tabcolsep}{4pt}
  \renewcommand{\arraystretch}{1.0}
  \begin{tabularx}{\textwidth}{@{}l r r X@{}}
    \toprule
    ID & Judge avg & Covered & Uncovered role evidence \\
    \midrule
    en\_021 & 99.5 & 5/9 (55.6\%) & Mafia secrets, demon names, and the Fate \& Herbs identity are never probed; the transcript drifts into an old-photo side thread. \\
    en\_007 & 94.9 & 6/12 (50.0\%) & The Thornfield maid backstory, inherited-estate relationship, and abandonment motive are not elicited across the 102-message dialogue. \\
    en\_012 & 86.8 & 6/14 (42.9\%) & The invisible-nerd background, game-designer identity, and reunion premise are mostly absent; the dialogue repeats emotional confirmation. \\
    en\_008 & 83.8 & 5/12 (41.7\%) & The Yorkshire sanatorium rules, full-name address habit, solitary dining, and mysterious past do not surface in the transcript. \\
    zh\_049 & 75.4 & 5/20 (25.0\%) & Swallowed Star timeline events and named world facts are skipped while the dialogue spends 102 turns in generic action scenes. \\
    \bottomrule
  \end{tabularx}
  \caption{Representative case-level evidence gaps in released free-chat transcripts. Judge avg is from MiniMax-M2-her run\_1. Coverage is computed after the full 102-message transcript. Across these five cases, only 27/67 checklist items are covered; 40/67 items (59.7\%) remain uncovered.}
  \label{tab:exp-case-evidence-gap}
\end{table}

Table~\ref{tab:exp-case-evidence-gap} shows that even when original judge scores are high, free-chat transcripts may not trigger core role facts, motivations, or world-state constraints. Across the five representative cases, 40/67 items remain uncovered (59.7\%). The core risk is that the dialogue evidence underlying those scores has not systematically reached role requirements.

\subsection{Ablation Study: Benchmark Protocol and Stability}
\label{sec:exp-ablation}

\subsubsection{Fixed First Message Pollution}

This ablation is a protocol hygiene check rather than a model-comparison result: it tests whether a fixed opening pollutes checklist evidence. Early \tracebench{} inserted a fixed first user message into the dialogue history as a seed opening, following the construction style of earlier benchmarks such as the MiniMax Role-play Benchmark. We compare the FixFirstMessage and UserAgentFirst protocols over 5 models $\times$ 200 cases, for 1,000 case-runs in total. This design also introduces a confound beyond evidence attribution: the fixed opening becomes an in-context example for the target model, so later replies may reflect adaptation to a benchmark-provided dialogue history rather than unseeded free-dialogue behavior.

\begin{table}[!htbp]
  \centering
  \captionsetup{font=footnotesize,skip=3pt}
  \small
  \setlength{\tabcolsep}{4pt}
  \renewcommand{\arraystretch}{1.0}
  \begin{tabular*}{\columnwidth}{@{\extracolsep{\fill}}lrr@{}}
    \toprule
    Metric & FixFirstMessage & UserAgentFirst \\
    \midrule
    case-runs & 1,000 & 1,000 \\
    checklist items & 27,490 & 27,490 \\
    pending$\rightarrow$resolved & 3,955 & 2,646 \\
    resolved / checklist items & 14.4\% & 9.6\% \\
    FM-involved resolved items & 1,130 & 0 \\
    FM-only contamination & 758 & 0 \\
    case-runs with FM contamination & 447 (44.7\%) & 0 \\
    \bottomrule
  \end{tabular*}
  \caption{First-round checklist states under FixFirstMessage and UserAgentFirst. The unit is the checklist state after the first evaluation of each case: all items are pending at round 0, and after the first round each item is recorded as completed, failed, or pending. \textit{pending$\rightarrow$resolved} counts checklist items that move from pending to completed or failed in the first round. \textit{resolved / checklist items} is the first-round resolved ratio. \textit{FM-involved resolved items} counts resolved items whose evidence matches the fixed roleplay opening, including mixed matches with the roleplay model reply. \textit{FM-only contamination} counts resolved items that match only the fixed roleplay opening and not the roleplay model reply. \textit{case-runs with FM contamination} counts cases with at least one FM-involved resolved item in the first round. Under FixFirstMessage, 44.7\% of case-runs are already contaminated in the first round; under UserAgentFirst, FM involvement is 0.}
  \label{tab:exp-first-message-states}
\end{table}

\begin{figure}[!htbp]
  \centering
  \includegraphics[width=0.85\columnwidth]{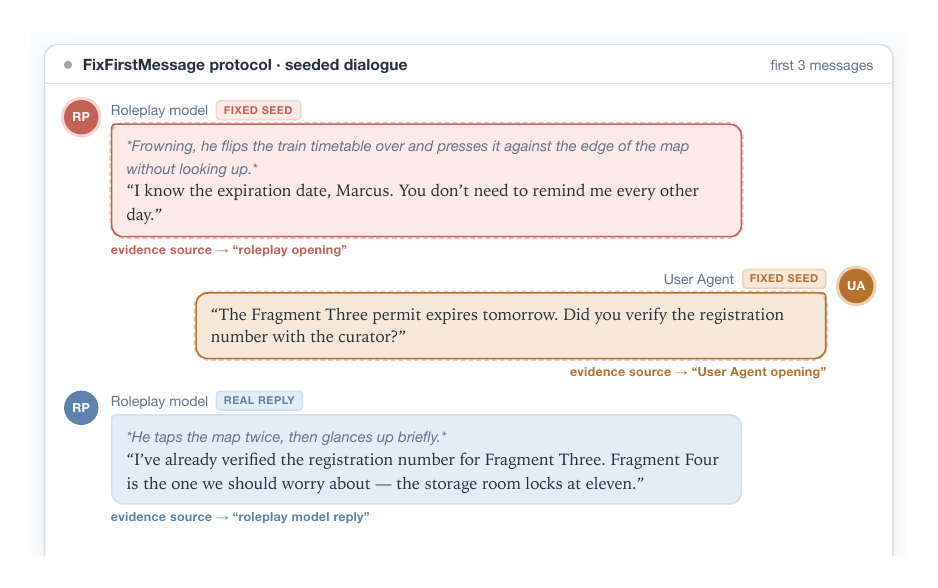}
  \caption{The FixFirstMessage seeded dialogue (first three messages). The two red-bordered messages are fixed benchmark-provided content: a roleplay opening and a User Agent opening. The roleplay model reply (blue) is the target model's real behavior; subsequent free dialogue is omitted. The \textit{Source} column in Table~\ref{tab:exp-first-message-sources} traces each resolved checklist item's evidence back to which of these seeded sources it matches.}
  \label{fig:seeded-dialogue}
\end{figure}

\begin{table}[!htbp]
  \centering
  \captionsetup{font=footnotesize,skip=3pt}
  \small
  \setlength{\tabcolsep}{4pt}
  \renewcommand{\arraystretch}{1.0}
  \begin{tabularx}{\columnwidth}{@{}l r X@{}}
    \toprule
    Source & Resolved & Meaning \\
    \midrule
    \textcolor{fmsrc}{Pure roleplay opening} & 758 & Matches only the fixed roleplay opening \\
    \textcolor{fmsrc}{Roleplay opening} + \textcolor{rpmodel}{roleplay model reply} & 372 & Matches both sources \\
    \textcolor{rpmodel}{Pure roleplay model reply} & 1,984 & Normal roleplay model behavior \\
    \textcolor{uaopen}{User Agent opening} & 22 & Matches only the fixed User Agent opening \\
    \textcolor{uaopen}{User Agent opening} + \textcolor{rpmodel}{roleplay model reply} & 113 & Mixed source \\
    \textcolor{unattr}{Unattributed} & 706 & Matches neither source \\
    \bottomrule
  \end{tabularx}
  \caption{Evidence-source breakdown for the first FixFirstMessage round. This table traces the 3,955 \textit{pending$\rightarrow$resolved} items and divides them into six categories. The 758 items in \textit{Pure roleplay opening} come entirely from the fixed opening: these items are judged completed or failed before the agent sends any user message, so their ``evidence'' comes from benchmark-provided content rather than roleplay model behavior. The 706 unattributed items account for $706/3{,}955=17.9\%$, indicating that the fixed opening makes the agent more likely to mark items complete without specific evidence.}
  \label{tab:exp-first-message-sources}
\end{table}

\FloatBarrier

Taken together, the fixed roleplay opening pollutes checklist evaluation in two ways. Directly, $1{,}130/3{,}955=28.6\%$ of first-round resolved items cite the fixed opening, and $758/3{,}955=19.2\%$ depend on it entirely (Tables~\ref{tab:exp-first-message-states}--\ref{tab:exp-first-message-sources}, Figure~\ref{fig:seeded-dialogue}). Indirectly, $706/3{,}955=17.9\%$ of resolutions are unattributed, suggesting the seeded opening also lowers the evidence threshold for marking items complete. Under UserAgentFirst, FM involvement drops to 0, removing both forms of contamination.

The final benchmark protocol therefore adopts UserAgentFirst: the fixed first user message is removed and the User Agent initiates the first turn by itself, which also removes the in-context-learning confound introduced by a benchmark-provided opening, so the first observable roleplay response no longer depends on a fixed dialogue seed. All subsequent experiments run under this protocol unless otherwise stated.

\FloatBarrier

\subsubsection{Repeated-run Stability}

We test the most direct form of reproducibility: when the roleplay model, benchmark, and configuration are fixed, do independent repeated runs change scores and rankings? Six roleplay models are each evaluated with 3 independent runs $\times$ 200 cases. Each run uses a different random seed. The roleplay model temperature is fixed at 0.8, and the User Agent temperature is fixed at 0.6. The dialogue-guiding ability comes entirely from the User Agent's checklist-driven decisions.

\begin{table}[!htbp]
  \centering
  \captionsetup{font=footnotesize,skip=3pt}
  \small
  \setlength{\tabcolsep}{5pt}
  \renewcommand{\arraystretch}{1.0}
  \begin{tabular*}{0.82\textwidth}{@{\extracolsep{\fill}}lcccc@{}}
    \toprule
    Model & R1 & R2 & R3 & $\pm$Std. \\
    \midrule
    Qwen3.5-27B & 92.11 (\#1) & 92.02 (\#1) & 91.75 (\#1) & $\pm$0.19 \\
    HER-32B & 89.27 (\#2) & 89.68 (\#2) & 89.11 (\#2) & $\pm$0.30 \\
    CoSER-Llama-3.1-70B & 86.32 (\#3) & 86.67 (\#3) & 86.38 (\#3) & $\pm$0.19 \\
    Ministral-3-14B & 79.56 (\#4) & 79.25 (\#4) & 78.61 (\#4) & $\pm$0.48 \\
    Hermes-4-14B & 74.14 (\#5) & 75.74 (\#5) & 76.80 (\#5) & $\pm$1.34 \\
    CoSER-Llama-3.1-8B & 63.36 (\#6) & 64.31 (\#6) & 63.36 (\#6) & $\pm$0.55 \\
    \bottomrule
  \end{tabular*}
  \caption{Repeated-run stability across six roleplay models. Each model has 3 runs $\times$ 200 cases. Scores are Overall scores, with ranks shown in parentheses.}
  \label{tab:exp-repeated-run}
\end{table}

Table~\ref{tab:exp-repeated-run} shows that the three repeated runs of all six models vary only within an acceptable range. The largest standard deviation is $\pm$1.34 for Hermes-4-14B, and all other models stay below $\pm$0.60. Under the same benchmark protocol and model configuration, repeated runs yield similar Overall scores and preserve the main model ranking.

\subsubsection{User Agent Model Replacement}

Repeated runs test whether randomness under the same User Agent changes conclusions. We further test whether results depend on a particular User Agent: six roleplay models are re-evaluated under three User Agents that can stably execute the \tracebench{} protocol, namely Qwen3.6-27B, Qwen3.5-27B, and Gemma-4-31B. A qualified User Agent is one that can maintain checklist state, generate natural user utterances, and end evaluation through the conversation-finish tool within a reasonable number of turns.

\begin{table}[!htbp]
  \centering
  \captionsetup{font=footnotesize,skip=3pt}
  \small
  \setlength{\tabcolsep}{4pt}
  \renewcommand{\arraystretch}{1.0}
  \begin{tabular*}{0.9\textwidth}{@{\extracolsep{\fill}}lrrr@{}}
    \toprule
    Roleplay model & Gemma-4-31B & Qwen3.5-27B & Qwen3.6-27B \\
    \midrule
    Qwen3.5-27B & 93.37 (\#1) & 92.70 (\#1) & 92.13 (\#1) \\
    HER-32B & 91.45 (\#2) & 89.82 (\#2) & 88.73 (\#2) \\
    CoSER-Llama-3.1-70B & 88.96 (\#3) & 85.51 (\#3) & 86.02 (\#3) \\
    Ministral-3-14B & 83.62 (\#4) & 81.57 (\#4) & 77.84 (\#4) \\
    Hermes-4-14B & 76.40 (\#5) & 75.65 (\#5) & 74.89 (\#5) \\
    CoSER-Llama-3.1-8B & 64.43 (\#6) & 65.14 (\#6) & 62.48 (\#6) \\
    \bottomrule
  \end{tabular*}
  \caption{User Agent replacement across six roleplay models. Scores are Overall scores; ranks are shown in parentheses.}
  \label{tab:exp-ua-replacement}
\end{table}

The three qualified User Agents produce exactly the same Overall ordering, with no rank changes.

\subsection{Full 26-Model Benchmark Application}
\label{sec:exp-comprehensive-comparison}

After establishing coverage validity and protocol stability, we apply \tracebench{} to 26 target models and report the current reference leaderboard. The role of this leaderboard is simple: it places 26 target models under the same 200 cases, 5,498 checklist items, and fixed scoring contract, and shows their current ranking and component capability profile. The model set covers closed API models, open general models, and open roleplay-tuned models; type-specific rankings and additional breakdowns for the 78 CharacterEval-derived cases and the 122 scenario-generated cases are reported in Appendix~\ref{sec:app-leaderboard-breakdowns}. The leaderboard is application evidence; the validity argument comes from the coverage and stability experiments above, not from the ranking table alone.

The full leaderboard is based on 200 cases and 5,498 checklist items. Overall is computed by the weighted composite in the Experimental Setup. The main benchmark run reaches 100.00\% Covered Rate for all models. We therefore omit Covered Rate from the leaderboard and keep completed-to-failed checklist flips (C$\rightarrow$F) as a diagnostic column after the scoring dimensions; C$\rightarrow$F is not used in the Overall score.

\begin{table}[!htbp]
  \centering
  \captionsetup{font=footnotesize,skip=3pt}
  \footnotesize
  \setlength{\tabcolsep}{4pt}
  \renewcommand{\arraystretch}{0.92}
  \begin{tabular}{@{}r p{0.30\textwidth} l r r r r r r r@{}}
    \toprule
    Rank & Model & Type & Overall & CC & STM & Div. & LQ & Len. & C$\rightarrow$F $\downarrow$ \\
    \midrule
    1 & Gemma-4-31B & Open & 96.02 & 95.46 & 90.50 & 93.49 & 97.66 & 98.49 & 4 \\
    2 & GPT-5.5 & API & 95.30 & 94.42 & 92.00 & 92.58 & 97.87 & 96.54 & 5 \\
    3 & Claude-Sonnet-4.6 & API & 94.72 & 94.68 & 86.00 & 92.22 & 98.16 & 93.70 & 7 \\
    4 & Mimo-v2.5 & API & 94.43 & 93.94 & 91.00 & 86.62 & 99.03 & 94.59 & 5 \\
    5 & Kimi-K2.5 & API & 94.31 & 94.25 & 87.50 & 88.70 & 98.16 & 94.06 & 4 \\
    6 & Mimo-v2.5-Pro & API & 93.98 & 94.95 & 90.00 & 87.59 & 98.73 & 88.75 & 3 \\
    7 & DeepSeek-V4-Pro & API & 93.96 & 93.56 & 85.00 & 95.43 & 97.41 & 91.42 & 8 \\
    8 & MiniMax-M2.7 & API & 93.88 & 94.00 & 88.50 & 87.82 & 96.53 & 94.93 & 12 \\
    9 & Ditto-8B & RP & 93.47 & 92.91 & 93.00 & 82.76 & 96.52 & 97.39 & 9 \\
    10 & DeepSeek-V4-Flash & API & 92.79 & 91.81 & 87.00 & 91.79 & 98.58 & 88.70 & 16 \\
    11 & GLM-5 & API & 92.28 & 93.94 & 87.00 & 88.24 & 95.51 & 86.38 & 9 \\
    12 & Qwen3.5-27B & Open & 92.10 & 94.53 & 85.00 & 89.48 & 96.59 & 81.41 & 6 \\
    13 & M2-HER & RP & 91.03 & 87.82 & 79.00 & 90.23 & 97.26 & 94.79 & 18 \\
    14 & Qwen3-14B & Open & 90.62 & 88.36 & 84.00 & 74.76 & 96.98 & 99.60 & 12 \\
    15 & HER-32B & RP & 89.12 & 85.21 & 73.50 & 92.23 & 96.93 & 90.95 & 49 \\
    16 & Qwen3-8B & Open & 88.43 & 85.44 & 85.50 & 66.45 & 97.27 & 98.32 & 28 \\
    17 & CoSER-Llama-3.1-70B & RP & 86.04 & 83.19 & 81.50 & 81.72 & 94.84 & 84.32 & 26 \\
    18 & Mistral-Small-3.2-24B & Open & 85.71 & 87.57 & 84.50 & 56.69 & 90.77 & 91.44 & 28 \\
    19 & Ministral-3-14B & Open & 78.55 & 87.82 & 74.00 & 90.11 & 66.46 & 64.68 & 44 \\
    20 & Hermes-4-14B & RP & 76.01 & 75.46 & 65.50 & 60.47 & 84.91 & 76.70 & 37 \\
    21 & L3-8B-Stheno-v3.2 & RP & 74.34 & 81.11 & 85.43 & 67.76 & 65.42 & 69.57 & 58 \\
    22 & Qwen2.5-7B-RoleMRC-DPO & RP & 71.97 & 69.28 & 66.50 & 56.85 & 74.82 & 87.21 & 40 \\
    23 & Qwen3.5-2B & Open & 67.22 & 71.19 & 63.50 & 78.61 & 59.18 & 62.38 & 73 \\
    24 & Crab & RP & 66.48 & 56.60 & 64.00 & 74.06 & 70.11 & 85.87 & 73 \\
    25 & Llama-3.1-8B-RoleMRC-DPO & RP & 65.92 & 68.60 & 78.50 & 41.05 & 62.77 & 75.54 & 59 \\
    26 & CoSER-Llama-3.1-8B & RP & 64.52 & 53.47 & 58.50 & 71.49 & 74.82 & 77.85 & 48 \\
    \bottomrule
  \end{tabular}
  \caption{Full 26-model leaderboard. Type labels are descriptive: API = closed API model, Open = open general model, and RP = open roleplay-tuned model. Results are based on 200 cases and 5,498 checklist items. Overall is computed as $0.45\times\mathrm{CC} + 0.05\times\mathrm{STM} + 0.10\times\mathrm{Diversity} + 0.25\times\mathrm{LQ} + 0.15\times\mathrm{Length}$, where LQ denotes Language Quality. C$\rightarrow$F is a diagnostic column and is not used in the Overall score; it counts completed-to-failed checklist flips.}
  \label{tab:exp-comprehensive-comparison}
\end{table}

\FloatBarrier

\subsection{Closed-Loop Benchmark Evolution}
\label{sec:exp-closed-loop-evolution}

We then test whether Closed-Loop Benchmark Evolution can turn observed effective verification methods into more targeted follow-up tests. The experiment proceeds in three steps. First, we evaluate the same 26 roleplay models on the base \tracebench{} benchmark and retain their checklist state traces and dialogue evidence. Second, we derive new verification flows from the failed traces in these runs, yielding an evolved benchmark. Third, we re-evaluate the same 26 models on the evolved benchmark and compare scores before and after evolution.

This experiment focuses on Character Consistency (CC), because CC directly measures whether role requirements are satisfied. The evolved benchmark preserves the role profiles, evaluated checklist requirements, and CC scoring denominator. The main change is how the User Agent organizes the conversation, follows up on boundaries, revisits earlier facts, and elicits conditional failures. A drop in CC therefore indicates that the same role requirements expose more model weaknesses under the evolved verification flows.

Results show that all 26 models obtain lower CC on the evolved benchmark, with an average drop of 8.48 points. This indicates that Closed-Loop Benchmark Evolution can distill verification methods that proved effective in the first-round evaluation into a more targeted stress-test protocol, enabling the User Agent to elicit checklist requirements from the role profile in a more natural and focused way. Table~\ref{tab:exp-closed-loop-evolution} reports the per-model comparison.

\begin{table}[H]
  \centering
  \captionsetup{font=footnotesize,skip=3pt}
  \small
  \setlength{\tabcolsep}{0pt}
  \renewcommand{\arraystretch}{0.92}
  \begin{tabular*}{0.80\textwidth}{@{\extracolsep{\fill}}l r r r@{}}
    \toprule
    Model & Base CC & Evolved CC & $\Delta$CC \\
    \midrule
    Gemma-4-31B & 95.46 & 90.73 & -4.73 \\
    GPT-5.5 & 94.42 & 90.78 & -3.64 \\
    Claude-Sonnet-4.6 & 94.68 & 89.28 & -5.40 \\
    Mimo-v2.5 & 93.94 & 86.00 & -7.94 \\
    Kimi-K2.5 & 94.25 & 87.88 & -6.37 \\
    Mimo-v2.5-Pro & 94.95 & 86.48 & -8.47 \\
    DeepSeek-V4-Pro & 93.56 & 86.43 & -7.13 \\
    MiniMax-M2.7 & 94.00 & 88.70 & -5.30 \\
    Ditto-8B & 92.91 & 86.69 & -6.22 \\
    DeepSeek-V4-Flash & 91.81 & 84.80 & -7.01 \\
    GLM-5 & 93.94 & 91.11 & -2.83 \\
    Qwen3.5-27B & 94.53 & 88.43 & -6.10 \\
    M2-HER & 87.82 & 80.32 & -7.50 \\
    Qwen3-14B & 88.36 & 79.58 & -8.78 \\
    HER-32B & 85.21 & 73.21 & -12.00 \\
    Qwen3-8B & 85.44 & 75.63 & -9.81 \\
    CoSER-Llama-3.1-70B & 83.19 & 74.44 & -8.75 \\
    Mistral-Small-3.2-24B & 87.57 & 78.94 & -8.63 \\
    Ministral-3-14B & 87.82 & 80.62 & -7.20 \\
    Hermes-4-14B & 75.46 & 59.16 & -16.30 \\
    L3-8B-Stheno-v3.2 & 81.11 & 73.22 & -7.89 \\
    Qwen2.5-7B-RoleMRC-DPO & 69.28 & 54.42 & -14.86 \\
    Qwen3.5-2B & 71.19 & 59.25 & -11.94 \\
    Crab & 56.60 & 47.99 & -8.61 \\
    Llama-3.1-8B-RoleMRC-DPO & 68.60 & 49.73 & -18.87 \\
    CoSER-Llama-3.1-8B & 53.47 & 45.16 & -8.31 \\
    \midrule
    Mean & 84.98 & 76.50 & -8.48 \\
    \bottomrule
  \end{tabular*}
  \caption{Closed-Loop Benchmark Evolution comparison on Character Consistency (CC). Base and evolved runs use the same role profiles, evaluated checklist requirements, and CC scoring denominator. The evolved run only changes the User Agent verification flows induced from failed traces. $\Delta$CC is Evolved CC $-$ Base CC.}
  \label{tab:exp-closed-loop-evolution}
\end{table}
\FloatBarrier

\subsection{Human Checklist Agreement Audit}
\label{sec:exp-human-checklist-agreement}

As an additional validation, we conduct a human checklist agreement audit to test whether \tracebench{} checklist-state tracking is reliable. We randomly sample 600 non-overlapping checklist items from the benchmark trajectories of the 26 evaluated models, hide model identities and the User Agent's original state judgments, and ask seven human annotators to independently review the corresponding role profile, full dialogue transcript, and checklist item. Each annotator labels the final state of each checklist item as either \statelabel{completed} or \statelabel{failed}.

We use the human majority vote as the reference label and compute agreement between the User Agent's original judgment and this human reference label. We also report Fleiss' $\kappa$ to measure inter-annotator agreement among the seven annotators after correcting for chance agreement. The User Agent agrees with the human majority label on 558 of 600 checklist items, corresponding to 93.00\% agreement. The annotators achieve Fleiss' $\kappa=0.7255$. Appendix~\ref{sec:app-human-audit} reports the annotation protocol and vote-strength distribution.

%% file: sections/06_limitations.tex
\section{Limitations}
\label{sec:limitations}

Checklist completion is well suited to requirements that can be grounded in explicit role constraints, world rules, task hooks, and observable dialogue evidence. It is less complete as a measure of subjective experience. Naturalness, emotional nuance, character charm, pacing, immersion, and aesthetic fit are difficult to reduce to the same kind of discrete checklist state. A model may satisfy many checklist requirements while still producing a flat or unconvincing interaction.

\tracebench{} also depends on the quality of checklist construction and on the User Agent's ability to follow the protocol. A noisy role profile, an incomplete checklist, or a weak User Agent can reduce coverage and evidence quality. We mitigate these risks through automatic validation, manual review, and User Agent qualification tests, but they remain practical constraints for future benchmark expansion.

The current framework uses per-turn LLM-judge scores as a supplementary signal for subjective quality, but this signal is still mechanical. Per-turn judgment can fragment a trajectory, miss the cumulative effect of earlier choices, and inherit the judge's own prompt sensitivity and model preferences. Future work should develop more dynamic, trajectory-level methods for subjective quality evaluation.

%% file: sections/07_conclusion.tex
\section{Conclusion}
\label{sec:conclusion}

\tracebench{} moves roleplay evaluation from black-box holistic scoring toward traceable interactive verification. By converting role profiles into fixed checklist requirements and using a User Agent to actively elicit and record those requirements in natural multi-turn dialogue, \tracebench{} makes model scores explainable through concrete checklist states and dialogue evidence. The experiments show that this protocol improves requirement coverage and preserves stable rankings in repeated-run and User-Agent-replacement tests; the 26-model leaderboard demonstrates its fine-grained diagnostic value for current roleplay models. Closed-Loop Benchmark Evolution further distills verification methods that prove effective in evaluation traces into finer-grained verification flows, extending \tracebench{} from a one-shot benchmark into a maintainable and iterative roleplay evaluation infrastructure.

%% file: sections/08_appendix.tex
\section{Example Benchmark Case and Verification Flow Evolution}
\label{sec:app-case-example}

This appendix gives a one-page example of the case format used by TRACE Bench. It shows one scenario-generated English case, a sequential checklist excerpt, and two checklist-level examples of how Closed-Loop Benchmark Evolution rewrites verification flow. The gray boxes preserve the original case structure and use ellipses to omit repeated or less relevant lines.

\subsection{Case Profile and Checklist Excerpt}

\begingroup
\setlength{\fboxsep}{4.5pt}
\noindent\fcolorbox{black!20}{black!4}{%
\begin{minipage}{0.94\textwidth}
\scriptsize
\renewcommand{\arraystretch}{0.98}
\begin{tabular}{@{}p{0.455\textwidth}@{\hspace{0.035\textwidth}}p{0.455\textwidth}@{}}
\textbf{Role profiles} & \textbf{Checklist items and base verification flows} \\
\midrule
\begin{minipage}[t]{\linewidth}
\raggedright
\textbf{Target role profile.}\par
\textbf{Character Name and Profile.} Mateo Vilar, 45, Senior Accounting Director at the Port of Barcelona; nearly two decades in the container-yard office; red line: clean accounts and compliant documents.\par
\textbf{Speaking Style.} Spanish-English port terminology, precise wording, moderate pace, and phrases such as ``mira'', ``escucha'', ``el tema es que'', B/L, DUA, AEAT, and UCC.\par
\textbf{Personality Traits.} Number-sensitive, cautious but not cold, politically neutral, strict about work/family boundaries, and occasionally dryly humorous under absurd pressure.\par
\textbf{Current Relationship with the Player.} Lucía Ortega is a distant cousin and freight-forwarding sales representative who asks him to release goods first and handle tax later.\par
\textbf{Specific Behavioral Patterns.} Mateo checks documents or screens, explains the regulatory obstacle, offers compliant alternatives, and calmly closes the topic under repeated pressure.\par
\textbf{Abilities and Knowledge Boundaries.} He knows UCC, customs clearance, container-yard protocol, port labor agreements, and AEAT collaboration; he avoids legal advice, politics, and duties outside port accounting.\par
\textbf{Dialogue Examples.} The original case includes release-first requests, family pressure, compliant alternatives, and continued pressure \ldots{}\par
\medskip
\textbf{User role profile.}\par
\textbf{User Identity.} Lucía Ortega, early 30s, sales representative at a Barcelona freight-forwarding agency and Mateo's distant cousin.\par
\textbf{Speaking Style.} Short, choppy, direct chat messages, usually 4--14 words, without narration, long explanation, or formal tone.\par
\textbf{Relationship with Target.} She knows the request is difficult but hopes family ties will make him bend the rules.\par
\textbf{Current Scene.} Container-yard office; she walks in with her voice lowered while machinery is loud outside and the office is quiet.\par
\textbf{Purpose / Goal.} Get Mateo's signature and release the shipment documents today, leaving the tax issue for later.\par
\end{minipage}
&
\begin{minipage}[t]{\linewidth}
\raggedright
\textbf{1.} The target should self-identify as Mateo Vilar, Senior Accounting Director of the Port de Barcelona. \emph{Flow:} request signatory confirmation and check exact name/title.\par
\textbf{2.} The target should display nearly twenty years of container-yard office seniority. \emph{Flow:} probe veteran familiarity with the yard and procedures.\par
\textbf{3.} The target should notice account discrepancies, even at decimal-point level. \emph{Flow:} alter a figure under pressure and check whether he flags it.\par
\textbf{4.} The target should be cautious but not cold: listen, explain, and suggest compliant routes without granting a personal favor. \emph{Flow:} ask for a rule-bending workaround.\par
\textbf{5.} The target should avoid taking sides among unions, freight forwarders, and shipping companies. \emph{Flow:} frame the request as a favor backed by outside influence.\par
\textbf{6.} The target should keep work and family relationships separate. \emph{Flow:} use family pressure and check whether he preserves the professional boundary.\par
\textbf{7.} The target should recognize Lucía as a distant cousin rather than a close daily contact. \emph{Flow:} imply closeness and observe whether he corrects the relation.\par
\textbf{8.} The target should seriously explain why the request is impossible without becoming hostile. \emph{Flow:} escalate family appeals and urgent procedural demands.\par
\textbf{9.} When Lucía lowers her voice, the target should maintain normal volume. \emph{Flow:} begin with a lowered-voice request and observe response style.\par
\textbf{10.} The target should check documents or the screen before giving numbers or clauses. \emph{Flow:} ask a specific factual question under time pressure.\par
\textbf{\ldots{}} Items 11--33 continue with ``el tema es que'' explanations, cumulative pressure handling, compliant alternatives, family-topic deflection, customs-clearance red lines, Spanish-English style, UCC/AEAT knowledge, legal-advice boundaries, and cross-turn recall.\par
\end{minipage}
\end{tabular}
\end{minipage}%
}
\endgroup

\vspace{2pt}
\subsection{Checklist-Level Flow Evolution}

The same case also illustrates flow-level evolution: the role profile and checklist requirements stay fixed, while the User Agent's verification path becomes more targeted.

\begingroup
\setlength{\fboxsep}{4.5pt}
\noindent\fcolorbox{black!20}{black!4}{%
\begin{minipage}{0.94\textwidth}
\scriptsize
\renewcommand{\arraystretch}{1.0}
\begin{tabularx}{\textwidth}{@{}X@{\hspace{0.04\textwidth}}X@{}}
\textbf{Base verification flow} & \textbf{Evolved verification flow} \\
\midrule
\textbf{No release before clearance.} Repeatedly request release first and tax payment later; the target should hold the red line.
&
\textbf{No release before clearance.} Add urgency, financial stakes, and B/L details. Pass: cites the clearance/tax rule across turns. Fail: concedes or omits the constraint. \\
\addlinespace[2pt]
\textbf{Compliant alternatives.} After rejection, observe whether the target proactively offers aplazamiento or other compliant alternatives.
&
\textbf{Compliant alternatives.} Cycle urgent appeals while switching languages. Pass: proposes aplazamiento or formal review. Fail: becomes repetitive or omits alternatives.
\end{tabularx}
\end{minipage}%
}
\endgroup

\section{User Agent Tool Schema}
\label{sec:app-tool-schema}

The User Agent coordinates evaluation through two private tools. The checklist-update tool maintains checklist state, and the conversation-finish tool ends the conversation. Runtime schemas constrain both tools, and tool results are visible only to the User Agent. They never enter the public dialogue. At the start of each case, the User Agent receives a fixed prebuilt checklist and mainly uses update operations to revise existing items against new evidence. Add operations are reserved for occasional role-relevant test points that are absent from the original checklist. The conversation-finish tool may be called only when every checklist item has reached a terminal judgment with sufficient evidence.

\newcommand{\schematool}[1]{\multicolumn{4}{@{}l}{\textbf{#1}}\\[-1pt]}
\newcommand{\schemafield}[1]{\hspace{0.75em}\textit{#1}}
\newcommand{\schemavalue}[1]{\textit{#1}}
\begin{table}[H]
  \centering
  \scriptsize
  \setlength{\tabcolsep}{3pt}
  \renewcommand{\arraystretch}{0.9}
  \begin{tabularx}{\textwidth}{@{}p{3.25cm}p{2.15cm}cX@{}}
    \toprule
    \textbf{Field} & \textbf{Type} & \textbf{Req.} & \textbf{Description} \\
    \midrule
    \schematool{Checklist-update tool}
    \schemafield{id} & \schemavalue{string} & yes & Stable task identifier; updates must reuse existing ids \\
    \schemafield{operation} & \schemavalue{add}/\schemavalue{update} & -- & Modify an existing item or add an occasional role-relevant new item \\
    \schemafield{content} & \schemavalue{string} & -- & Requirement being tested \\
    \schemafield{status} & \schemavalue{string} & -- & One of \statelabel{pending}, \statelabel{in\_progress}, \statelabel{completed}, \statelabel{failed}, or \statelabel{abandoned} \\
    \schemafield{priority} & \schemavalue{high}/\schemavalue{medium}/\schemavalue{low} & -- & Priority for online scheduling \\
    \schemafield{evidence} & \schemavalue{string} & -- & Brief evidence supporting the judgment \\
    \schemafield{note} & \schemavalue{string} & -- & Staged observations, questions, or forward testing plan \\
    \schemafield{attempted} & \schemavalue{boolean} & -- & Whether this turn attempted to trigger the item \\
    \schemafield{attempt evidence} & \schemavalue{string} & -- & How this turn tested the item and the outcome \\
    \schemafield{reason} & \schemavalue{string} & -- & Explanation for this checklist update \\
    \addlinespace[1pt]
    \cmidrule(lr){1-4}
    \schematool{Conversation-finish tool}
    \schemafield{reason} & \schemavalue{string} & yes & Reason for ending the conversation \\
    \schemafield{summary} & \schemavalue{string} & -- & Summary of the evaluation dialogue \\
    \bottomrule
  \end{tabularx}
  \caption{Field specification for the User Agent tools.}
  \label{tab:app-tool-schema}
\end{table}

\section{Scoring Formulas and Definitions}
\label{sec:app-scoring}

Let $C$ denote the evaluated case set ($|C|=200$ for the full benchmark; subset leaderboards use the corresponding subset), and let $I_{\text{non-STM}}(C)$ be the set of all prebuilt checklist items in $C$ excluding the cross-turn memory probe item. The five scoring dimensions and the overall composite are defined as follows:

\begingroup
\small
\setlength{\jot}{1pt}
\[
\begin{aligned}
\text{CC} &= 100 \times \frac{|\{i \in I_{\text{non-STM}}(C) : \text{status}(i) = \statevalue{completed}\}|}{|I_{\text{non-STM}}(C)|}, \\
\text{STM} &= 100 \times \frac{|\{c \in C : \text{STM}(c) = \statevalue{completed}\}|}{|C|}, \\
\text{LQ} &= 100 \times \frac{1}{N_{\text{LQ}}} \sum_{r} \text{lq\_delta}(r), \\
\text{Diversity} &= 100 \times \frac{1}{N_{\text{div}}} \sum_{r} \text{diversity\_delta}(r), \\
\text{Length} &= 100 \times \frac{1}{N_{\text{len}}} \sum_{r} \text{length\_delta}(r), \\
\text{Overall} &= 0.45 \times \text{CC} + 0.05 \times \text{STM} + 0.10 \times \text{Diversity} + 0.25 \times \text{LQ} + 0.15 \times \text{Length}.
\end{aligned}
\]
\endgroup

CC measures whether the target model satisfies the concrete checklist requirements derived from the role profile. STM measures whether the model remembers user-injected external facts from the multi-turn dialogue. LQ calls a language-quality judge once for each scorable roleplay reply. Given the preceding user input, the judge marks whether the reply contains obvious fluency, grammar, usage, or internal-logic problems; good and bad judgments map to \(\text{lq\_delta}=1\) and \(0\). Diversity splits each reply into sentences, filters very short sentences, compares current sentences with previous target-model sentences using character-bigram Jaccard similarity, and maps the maximum similarity to \(\text{diversity\_delta}\in[0,1]\): similarity \(\leq 0.4\) receives 1, similarity \(\geq 0.6\) receives 0, and the interval between them is linearly interpolated. Length uses deterministic bounds. English-majority replies are measured by word count and receive \(\text{length\_delta}=1\) when they contain 4--80 words; Chinese or mixed replies are measured by CJK or non-space character count and receive 1 when they contain 15--150 characters. Replies outside the corresponding range receive 0. All component scores are reported on a 0--100 scale; the three reply-level auxiliary dimensions are normalized means over scorable replies.

\section{Coverage Truncation Curve}
\label{sec:app-truncation-curve}

Table~\ref{tab:app-truncation-curve} reports the complete truncation curve for MiniMax free-dialogue coverage. The main text reports only the key truncation points used for comparison with agentic dialogue; this table provides the complete truncation grid. Each transcript is truncated to the first $N$ messages and checklist states are recomputed. The truncation grid uses key quantiles of the agentic dialogue message-count distribution (min 13, P25=21, median 25, P75=33, P90=47, max 65), plus the full free-dialogue length of 102.

\begin{table}[H]
  \centering
  \small
  \setlength{\tabcolsep}{4pt}
  \renewcommand{\arraystretch}{1.08}
  \begin{tabular*}{\columnwidth}{@{\extracolsep{\fill}}rrrrr@{}}
    \toprule
    Truncation (msgs) & Completed & Failed & Uncovered & Coverage \\
    \midrule
    13 & 574 & 5 & 533 & 52.07\% \\
    21 & 612 & 7 & 493 & 55.67\% \\
    25 & 637 & 8 & 467 & 58.00\% \\
    33 & 668 & 9 & 435 & 60.88\% \\
    47 & 712 & 14 & 386 & 65.29\% \\
    65 & 750 & 18 & 344 & 69.06\% \\
    102 & 796 & 24 & 292 & 73.74\% \\
    \bottomrule
  \end{tabular*}
  \caption{Complete truncation curve for MiniMax free-dialogue coverage. Records = 95, total checklist items = 1,112. Coverage = (completed + failed) / total.}
  \label{tab:app-truncation-curve}
\end{table}

\section{Human Checklist Agreement Audit Details}
\label{sec:app-human-audit}

The main text reports the headline result of the human checklist agreement audit. This appendix records the annotation protocol. The audit randomly samples 600 non-overlapping checklist items from the benchmark trajectories of the 26 evaluated models. Each annotation unit contains the role profile, the full dialogue transcript, and the checklist item to be judged; the target model identity and the User Agent's original state judgment are hidden from annotators. Seven human annotators independently read each unit and assign one of two terminal-state labels: \statelabel{completed} or \statelabel{failed}. We use the seven-annotator majority vote as the human reference label and compute agreement between the User Agent's original judgment and that reference label. Fleiss' $\kappa$ is computed over the seven raw annotator labels.

\begin{table}[H]
  \centering
  \small
  \setlength{\tabcolsep}{4pt}
  \renewcommand{\arraystretch}{1.05}
  \begin{tabularx}{0.92\textwidth}{@{}lX@{}}
    \toprule
    Item & Setting \\
    \midrule
    Annotation unit & One checklist item with its corresponding role profile and full dialogue transcript \\
    Sample size & 600 non-overlapping checklist items \\
    Source & Benchmark trajectories from 26 target models \\
    Hidden information & Target model identity and original User Agent state judgment \\
    Annotators & 7 independent human annotators \\
    Label space & \statelabel{completed} / \statelabel{failed} \\
    Reference label & Human majority vote \\
    Reported metrics & User Agent agreement with human majority vote; Fleiss' $\kappa$ \\
    \bottomrule
  \end{tabularx}
  \caption{Annotation protocol for the human checklist agreement audit.}
  \label{tab:app-human-audit-protocol}
\end{table}

\section{Additional Leaderboard Breakdowns}
\label{sec:app-leaderboard-breakdowns}

This appendix reports subset leaderboards on the 78 CharacterEval-derived cases and the 122 scenario-generated cases. These two breakdown tables help inspect whether model rankings and capability profiles remain consistent across different data sources. Each table reports Overall and the five component dimensions: CC, STM, Diversity, LQ, and Length.

\begin{table}[H]
  \centering
  \footnotesize
  \setlength{\tabcolsep}{2pt}
  \renewcommand{\arraystretch}{0.9}
  \begin{tabular*}{\textwidth}{@{\extracolsep{\fill}}r p{0.30\textwidth} r r r r r r@{}}
    \toprule
    Rank & Model & Overall & CC & STM & Diversity & LQ & Length \\
    \midrule
    1 & Gemma-4-31B & 97.72 & 97.80 & 94.87 & 98.34 & 96.61 & 99.88 \\
    2 & Mimo-v2.5-Pro & 97.52 & 97.52 & 93.59 & 96.87 & 98.43 & 97.75 \\
    3 & Mimo-v2.5 & 97.48 & 96.80 & 94.87 & 97.75 & 98.98 & 97.72 \\
    4 & Claude-Sonnet-4.6 & 97.43 & 96.53 & 94.87 & 98.46 & 98.09 & 99.21 \\
    5 & GPT-5.5 & 96.91 & 96.17 & 93.59 & 99.43 & 97.64 & 97.31 \\
    6 & DeepSeek-V4-Flash & 96.67 & 95.75 & 88.46 & 99.20 & 98.21 & 97.91 \\
    7 & DeepSeek-V4-Pro & 96.51 & 95.88 & 84.62 & 98.78 & 97.58 & 99.05 \\
    8 & Qwen3.5-27B & 96.43 & 97.47 & 93.59 & 97.33 & 94.65 & 96.65 \\
    9 & MiniMax-M2.7 & 96.09 & 96.47 & 88.46 & 96.33 & 97.09 & 95.69 \\
    10 & Kimi-K2.5 & 95.90 & 95.79 & 89.74 & 96.73 & 97.63 & 94.83 \\
    11 & M2-HER & 95.55 & 94.23 & 80.77 & 98.62 & 99.30 & 96.17 \\
    12 & Ditto-8B & 95.09 & 95.36 & 98.72 & 90.60 & 95.18 & 95.93 \\
    13 & GLM-5 & 94.05 & 95.27 & 94.87 & 98.53 & 97.16 & 81.95 \\
    14 & Qwen3-14B & 92.95 & 92.49 & 88.46 & 84.12 & 94.41 & 99.30 \\
    15 & Qwen3-8B & 91.62 & 88.35 & 88.46 & 85.75 & 96.02 & 99.03 \\
    16 & HER-32B & 91.41 & 89.89 & 82.05 & 97.38 & 95.98 & 87.50 \\
    17 & CoSER-Llama-3.1-70B & 88.57 & 88.37 & 85.90 & 94.38 & 90.59 & 82.80 \\
    18 & Mistral-Small-3.2-24B & 87.53 & 87.64 & 88.46 & 81.30 & 84.31 & 96.40 \\
    19 & Ministral-3-14B & 77.67 & 90.06 & 76.92 & 92.88 & 52.86 & 71.95 \\
    20 & Hermes-4-14B & 76.69 & 75.24 & 67.95 & 71.14 & 80.53 & 81.25 \\
    21 & L3-8B-Stheno-v3.2 & 75.81 & 85.16 & 88.31 & 86.17 & 44.66 & 88.62 \\
    22 & Qwen3.5-2B & 72.40 & 71.29 & 61.54 & 97.45 & 52.50 & 95.84 \\
    23 & Llama-3.1-8B-RoleMRC-DPO & 69.34 & 69.06 & 82.05 & 64.56 & 57.68 & 88.56 \\
    24 & Qwen2.5-7B-RoleMRC-DPO & 68.24 & 66.32 & 65.38 & 66.21 & 61.54 & 87.50 \\
    25 & CoSER-Llama-3.1-8B & 64.10 & 57.58 & 60.26 & 85.47 & 65.72 & 67.99 \\
    26 & Crab & 60.63 & 53.61 & 61.54 & 76.77 & 56.19 & 78.06 \\
    \bottomrule
  \end{tabular*}
  \caption{Leaderboard on the 78 CharacterEval-derived cases.}
  \label{tab:app-leaderboard-ce}
\end{table}

\begin{table}[H]
  \centering
  \footnotesize
  \setlength{\tabcolsep}{2pt}
  \renewcommand{\arraystretch}{0.9}
  \begin{tabular*}{\textwidth}{@{\extracolsep{\fill}}r p{0.30\textwidth} r r r r r r@{}}
    \toprule
    Rank & Model & Overall & CC & STM & Diversity & LQ & Length \\
    \midrule
    1 & Gemma-4-31B & 94.93 & 93.96 & 87.70 & 90.39 & 98.34 & 97.60 \\
    2 & GPT-5.5 & 94.27 & 93.31 & 90.98 & 88.20 & 98.02 & 96.04 \\
    3 & Kimi-K2.5 & 93.29 & 93.27 & 86.07 & 83.57 & 98.50 & 93.57 \\
    4 & Claude-Sonnet-4.6 & 92.99 & 93.50 & 80.33 & 88.23 & 98.20 & 90.19 \\
    5 & Mimo-v2.5 & 92.48 & 92.12 & 88.52 & 79.50 & 99.06 & 92.59 \\
    6 & MiniMax-M2.7 & 92.47 & 92.42 & 88.52 & 82.39 & 96.18 & 94.44 \\
    7 & Ditto-8B & 92.43 & 91.34 & 89.34 & 77.74 & 97.37 & 98.32 \\
    8 & DeepSeek-V4-Pro & 92.33 & 92.07 & 85.25 & 93.28 & 97.30 & 86.54 \\
    9 & Mimo-v2.5-Pro & 91.72 & 93.30 & 87.70 & 81.66 & 98.92 & 83.00 \\
    10 & GLM-5 & 91.15 & 93.09 & 81.97 & 81.66 & 94.45 & 89.21 \\
    11 & DeepSeek-V4-Flash & 90.32 & 89.29 & 86.07 & 87.05 & 98.82 & 82.81 \\
    12 & Qwen3.5-27B & 89.32 & 92.64 & 79.51 & 84.47 & 97.83 & 71.68 \\
    13 & Qwen3-14B & 89.13 & 85.71 & 81.15 & 68.78 & 98.62 & 99.79 \\
    14 & M2-HER & 88.13 & 83.72 & 77.87 & 84.86 & 95.95 & 93.91 \\
    15 & HER-32B & 87.65 & 82.22 & 68.03 & 88.94 & 97.54 & 93.16 \\
    16 & Qwen3-8B & 86.40 & 83.59 & 83.61 & 54.11 & 98.07 & 97.86 \\
    17 & Mistral-Small-3.2-24B & 84.54 & 87.52 & 81.97 & 40.95 & 94.90 & 88.27 \\
    18 & CoSER-Llama-3.1-70B & 84.42 & 79.88 & 78.69 & 73.63 & 97.55 & 85.29 \\
    19 & Ministral-3-14B & 79.11 & 86.39 & 72.13 & 88.34 & 75.16 & 60.03 \\
    20 & Hermes-4-14B & 75.57 & 75.59 & 63.93 & 53.66 & 87.71 & 73.79 \\
    21 & Qwen2.5-7B-RoleMRC-DPO & 74.36 & 71.18 & 67.21 & 50.86 & 83.31 & 87.03 \\
    22 & L3-8B-Stheno-v3.2 & 73.41 & 78.55 & 83.61 & 55.98 & 78.70 & 57.39 \\
    23 & Crab & 70.23 & 58.52 & 65.57 & 72.33 & 79.02 & 90.87 \\
    24 & CoSER-Llama-3.1-8B & 64.79 & 50.84 & 57.38 & 62.56 & 80.64 & 84.16 \\
    25 & Qwen3.5-2B & 63.91 & 71.13 & 64.75 & 66.57 & 63.45 & 40.99 \\
    26 & Llama-3.1-8B-RoleMRC-DPO & 63.74 & 68.30 & 76.23 & 26.02 & 66.03 & 67.21 \\
    \bottomrule
  \end{tabular*}
  \caption{Leaderboard on the 122 scenario-generated cases.}
  \label{tab:app-leaderboard-nonce}
\end{table}